# A novel algorithm can generate data to train machine learning models in conditions of extreme scarcity of real world data

Olivier Niel, MD, PhD (1)(2) *

(1) Kannerklinik, 4 rue Barblé, L1210 Luxembourg, Luxembourg

(2) Faculty of Science, Technology and Medicine, University of Luxembourg, 2 Avenue de l'Université, L4365, Esch-sur-Alzette, Luxembourg

* Correspondence: o.r.p.niel@free.fr (preferred), olivier.niel@ext.uni.lu

**Abstract**

**Background:** Training machine learning models requires large datasets. However, collecting, curating, and operating large and complex sets of real world data poses several problems, in terms of costs, ethical and legal issues, and data availability. To overcome these difficulties, we propose a novel algorithm which can generate large artificial datasets to train machine learning models, even in conditions of extreme scarcity of real world data.

**Methods:** The data generation algorithm is based on a genetic algorithm, which mutates randomly generated datasets subsequently used for training a neural network. After training, the performance of the neural network on a batch of real world data is considered a surrogate for the fitness of the generated dataset used for its training. As selection pressure is applied to the population of generated datasets, unfit generated datasets are discarded, and the fitness of the fittest generated datasets increases throughout generations.

**Results:** The performance of the data generation algorithm was measured on the Iris dataset and on the Breast Cancer Wisconsin diagnostic dataset. In conditions of real world data abundance, mean accuracy of machine learning models trained on generated data was comparable to mean accuracy of machine learning models trained on real world data (0.956 in both cases on the Iris dataset, $p = 0.6996$, and 0.9377 versus 0.9472 on the Breast Cancer dataset, $p = 0.1189$). In conditions of simulated extreme scarcity of real world data, mean accuracy of machine learning models trained on generated data was significantly higher than mean accuracy of comparable machine learning models trained on scarce real world data (0.9533 versus 0.9067 on the Iris dataset, $p < 0.0001$, and 0.8692 versus 0.7701 on the Breast Cancer dataset, $p = 0.0091$).

**Conclusion:** Here we propose a novel algorithm, which can generate large artificial datasets to train machine learning models, in conditions of extreme scarcity of real world data, and also when cost or data sensitivity becomes an obstacle to the collection of large real world datasets.

# 1. Introduction

Artificial intelligence algorithms are now being used extensively in most scientific disciplines. In particular, machine learning models have proven remarkably effective at solving complex classification or regression problems [1]. It is noteworthy that machine learning models need intensive training before they can be used; effective training requires significant computational resources, and, for intrinsic reasons, extremely large datasets [1].

However, collecting, curating, and operating large and complex sets of real world data poses several problems. First, these operations have substantial material and human costs. As an example, in 2021, global spending on big data and analytics reached 241 billion US dollars, and should peak above 655 billion US dollars by 2029 [2]. Second, data collection often raises ethical and legal issues, in terms of privacy and data protection. Indeed, when the nature of data becomes sensitive, which is the case with most personal data like medical health records, complex and costly arrangements are required to comply with local regulations and legislations, such as the General Data Protection Regulation in European countries, and to ensure that data are processed legally and ethically [3]. Lastly, data of interest can be particularly difficult to collect; this is the case when data are stored in different repositories which are not interoperable, or, in the worst case scenario, when real world data are scarce, for instance in patients suffering from rare diseases.

In order to overcome some of the limitations and obstacles inherent to the management of real world data, synthetic data generation algorithms have been proposed. These algorithms can generate artificial data whose characteristics and distribution are supposed to mimic these of the real world data of interest. Synthetic data could therefore be particularly helpful to train machine learning models. However in practice, synthetic data suffer from several drawbacks, including lack of realism, bias, or privacy issues, as theses data are still intrinsically related to real world data [4]. Moreover, synthetic data generation algorithms, like generative adversarial networks [5] or variational auto encoders [6], still need to be trained on a large amounts of real world data before they can start to generate accurate synthetic data.

To this day, there is no effective algorithm able to generate data under stringent conditions of structure or scarcity of real world data.

Here we propose a novel algorithm, based on a genetic algorithm [7], which can generate large artificial datasets used to train machine learning models, even in conditions of extreme scarcity of real world data.

## 2. Material and methods

### 2.1. Implementation

#### 2.1.1. Algorithm implementation

The data generation algorithm was implemented in C/C++ using Code::Blocks 20.03 with GNU GCC compiler following the GNU GCC C++ 17 language standard.

The genetic algorithm randomly applied point mutations to data of candidate generated datasets.

The neural network used in the data generation process was a feedforward fully connected multi layer perceptron. Activation function was leaky rectified linear unit for inner layers, and softmax function for output layer. The architecture of the neural network was kept constant throughout experiments conducted on the same database.

#### 2.1.2. Machine learning model implementation

Neural networks were implemented in Python using IDLE 3.6, with calls to the TensorFlow 2 API.

The neural network was a feedforward fully connected multi layer perceptron. Activation function was rectified linear unit for inner layers, and softmax function for output layer. The architecture of the neural network was kept constant throughout experiments conducted on the same database.

#### 2.1.3. Overfitting

The occurrence of overfitting was monitored throughout experiments.

Overfitting was prevented using early stopping, L2 regularization, and dropout strategies whenever appropriate.

## 2.2. Datasets

### 2.2.1. Iris dataset

The Iris dataset is a validated dataset used for machine learning experiments [8]; it was first established by Fischer in 1936. The Iris dataset used in this article was obtained online from the UCI Machine Learning Repository in October 2022 (https://archive.ics.uci.edu/ml/datasets/iris), and slightly differs from Fischer's original dataset, as described on the dataset web page.

The Iris dataset comprises 150 instances, with 4 attributes, split in 3 classes, with no missing values.

### 2.2.2. Breast Cancer Wisconsin diagnostic dataset

The Breast Cancer Wisconsin diagnostic dataset is a validated dataset used for machine learning experiments [9, 10]. The dataset used in this article was obtained from the UCI Machine Learning Repository in December 2022 (https://archive.ics.uci.edu/ml/datasets/Breast+Cancer+Wisconsin+%28Diagnostic%29). The Breast Cancer Wisconsin diagnostic dataset comprises 569 instances, with 30 attributes, split in 2 classes, with no missing values.

## 2.3. Normalization of data

Whenever relevant, data were normalized using the min-max scaling function, where:

$$X' = \frac{(X - X_{min})}{(X_{max} - X_{min})}$$

## 2.4. Experiments

All experiments were performed on at least 3 different generated datasets and 3 different, randomly-sampled instances of each real world dataset. Real world data were randomly allocated to a training/testing batch or to a validating batch. The training/testing batch was used to estimate the fitness of generated datasets during data generation.

After data generation, the fittest generated dataset and the training/testing batch were used to train machine learning models, which performances were comparatively analyzed on the validating batch, as detailed below. All machine learning experiments were performed at least 5 times on each generated dataset and each instance of real world dataset. Performance extremums of machine learning models were discarded.

## 2.5. Statistics

### 2.5.1. Software

Statistical analyses and distribution plots were performed with GraphPad Prism 5.0.

### 2.5.2. Statistical tests

Median, mean, and percentiles were used to characterize distributions of datasets.

Unpaired t-test, paired t-test and/or Mann-Whitney tests with corresponding p values and 95% confidence intervals were used for computation and analysis of statistical significance.

Linear and nonlinear regression analyses, with computation of goodness of fit parameters including R-square were used for characterization of learning models.

### 2.5.3. Significance

P values of less than 0.05 were considered significant.

# 3. Results

## 3.1. Algorithm design

### 3.1.1. General principle

The data generation algorithm proposed in this article is based on a genetic algorithm, which mutates randomly generated datasets subsequently used for training a neural network. After training, the performance of the neural network on a batch of real world data, measured by its mean squared error, is considered a surrogate for the fitness of the generated dataset used for training the neural network. As selection pressure is applied to the population of generated datasets, unfit generated datasets are discarded, and the fitness of the fittest generated datasets increases throughout generations, until desired fitness is achieved.

More details on the design of the data generation algorithm are provided in Figure 1.

### 3.1.2. Theoretical approach

Let us consider a neural network $f'$ approximating a function $f : A \to B$ [11].
Let $\{X, Y\}$ be a training batch, with $X$ increasingly fit inputs, and with $Y = B$.
Let $\{A_1, B\}$ be a fit testing batch such that $A_1 \subset A$.
Let $S$ be actual outputs of $f'$ during the data generation procedure.

Training and testing constraints apply as per definition of mean squared error minimization:

By definition
$$\begin{cases} (Y - S_X)^2 \to 0 \\ (B - S_{A_1})^2 \to 0 \end{cases}$$

Therefore
$$\begin{cases} (Y - f'(X))^2 \to 0 \\ (B - f'(A_1))^2 \to 0 \end{cases}$$

We can derive
$$\begin{cases} (B - f'(X))^2 \to 0 \\ (B - f'(A_1))^2 \to 0 \end{cases}$$

Let $Fit(X) = \dfrac{1}{(B-f'(X))^2 + (B-f'(A_1))^2}$ be the fitness of $X$.

Let $R \sim U(A)$ be a uniform random variable taking values $r$ in $A$.

By definition, $X$ is randomly initialized such that $\forall\, x_i \in X, x_i = r_i$.

Therefore, initial conditions on $X$ are $\begin{cases} X \subset A, \text{ with no significant order on } X \\ Fit(X) = 0 + \epsilon \end{cases}$

Let $GA$ be a genetic algorithm performing point mutations on $X$.

By definition, at generation $k$, $GA$ verifies $\begin{cases} \forall\, x_{k,i} \in X_k, P(x_{k,i} = r_{k,i}) = p, \text{ with } p \in\, ]0;1[ \\ Fit(X_k) > Fit(X_{k-1}) \end{cases}$ [12]

Therefore, at the end of data generation, $X$ verifies $\begin{cases} X \subset A, \text{ with at least partial order on } X \\ Fit(X) \gg 0 \end{cases}$

Of note, the algorithm design implies no condition on $X \cap A_1$.

The consequences of this particular algorithm design, including inherent generalization properties, are described in more details in the discussion section of this article.

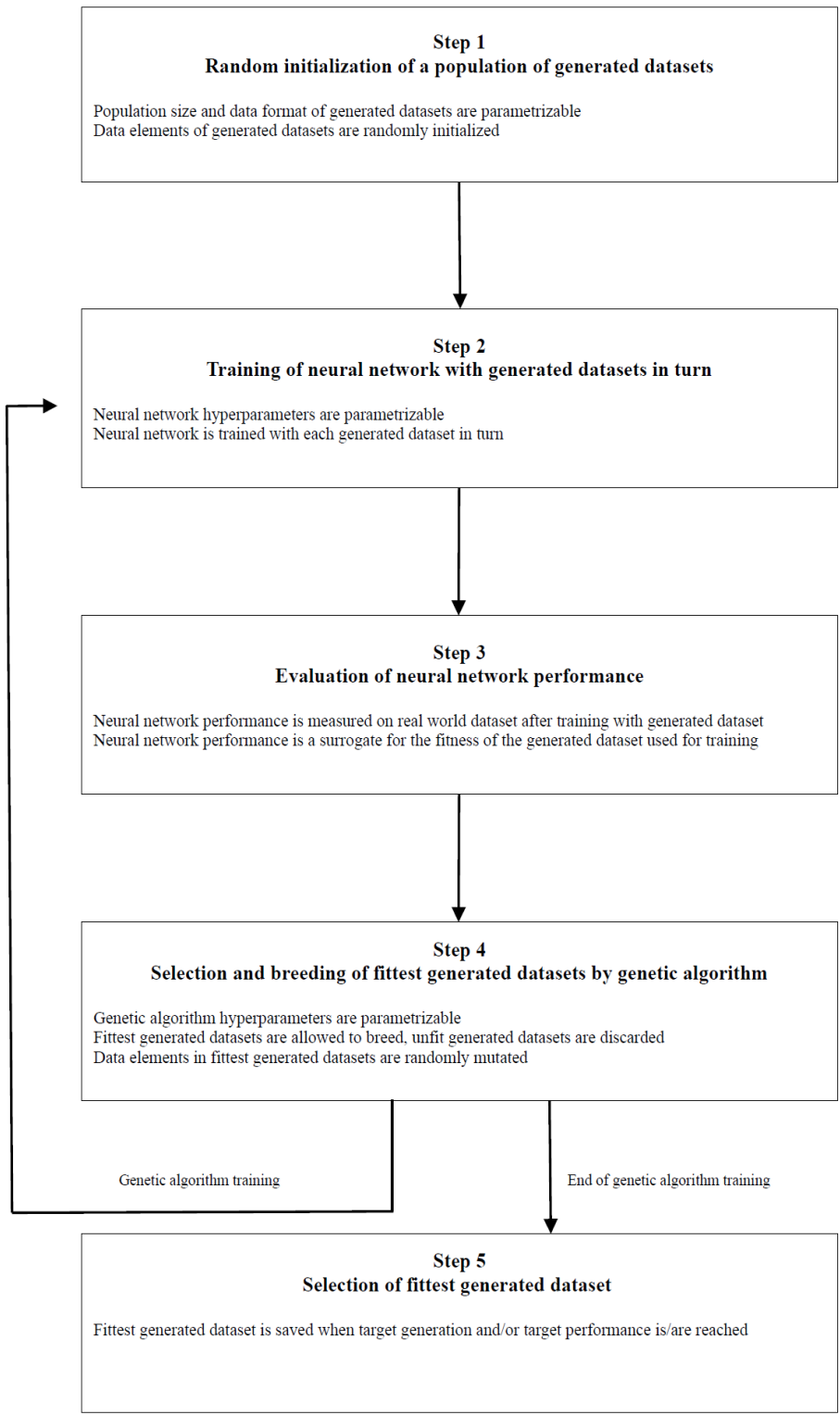

**Figure 1:** Algorithm design

## 3.2. Experimental workflow

The experimental workflow is described in Figure 2. Briefly, real world data were randomly split into a training/testing batch and a validating batch, in variable proportions. The training/testing batch was used to estimate the fitness of generated datasets during data generation.

After data generation, the fittest generated dataset and the training/testing batch were used to train machine learning models, which performances were comparatively analyzed on the validating batch in order to estimate the quality of the generated data.

## 3.3. Algorithm performance in conditions of abundance of real world data

At this stage, real world data were considered abundant. Fittest generated dataset and large real world data training/testing batch were used to train neural networks. The performances of the machine learning models were comparatively analyzed on real world data validating batch. Detailed results are provided in Table 1A and 1B.

### 3.3.1 Performance on the Iris dataset

The performance of the data generation algorithm was first assessed on the real world Iris dataset. For each experiment, the Iris dataset was randomly split into a training/testing batch, and validating batch. At this stage, real world data were considered abundant. The training/testing batch comprised 70% of the instances; the validating batch comprised 30% of the instances. A total of 150 instances were generated by the algorithm in each generated dataset, equally distributed between classes.

On the Iris dataset, the mean accuracy of the neural networks was comparable between models trained on generated datasets, and models trained on training/testing batches (0.956 in both cases, $p = 0.6996$).

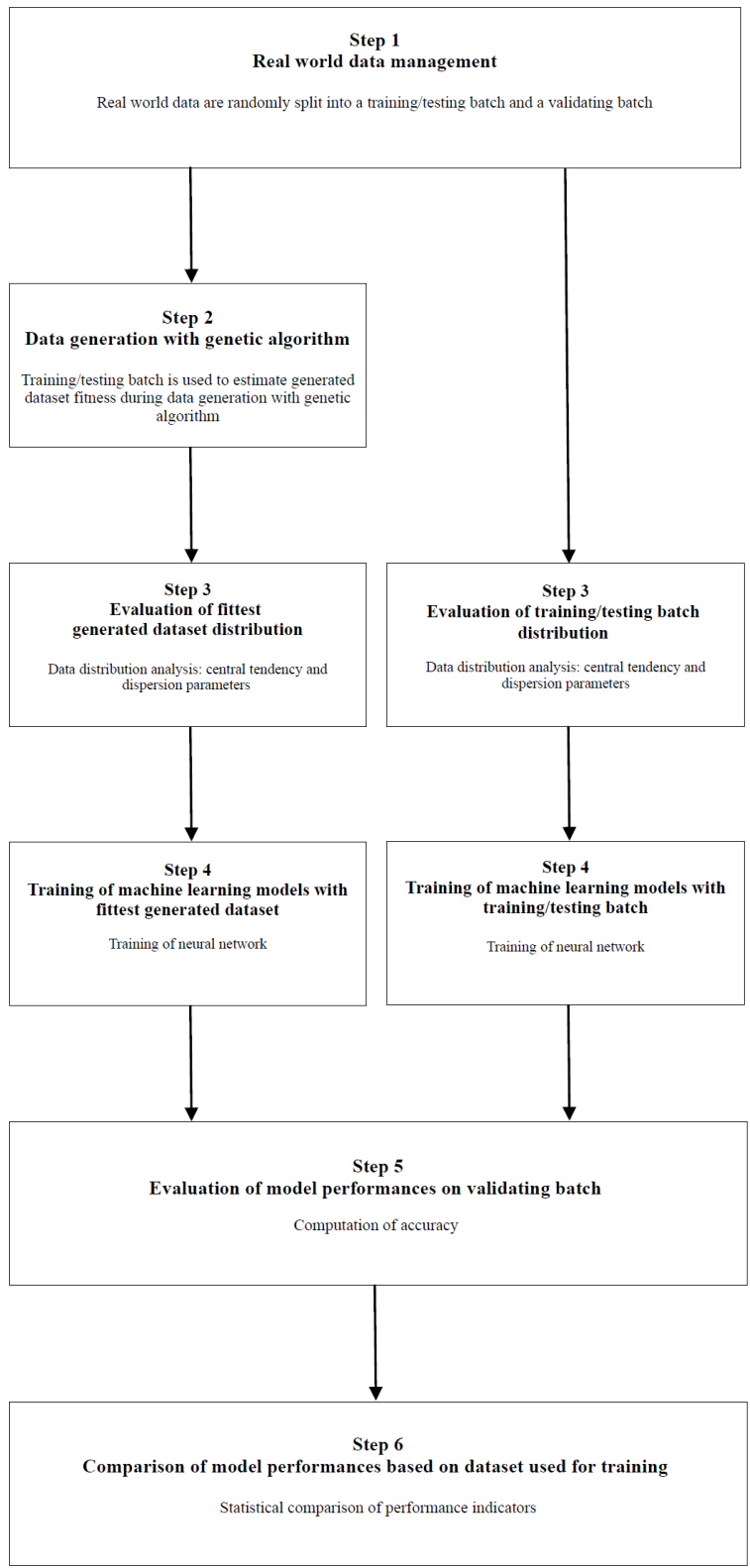

**Figure 2:** Experimental workflow

*3.3.2 Performance on the Breast Cancer Wisconsin diagnostic dataset*

The performance of the data generation algorithm was then assessed on the real world Breast Cancer Wisconsin diagnostic dataset. For each experiment, the dataset was randomly split into a training/testing batch, and validating batch. At this stage, real world data were considered abundant. The training/testing batch comprised 80% of the instances; the validating batch comprised 20% of the instances. A total of 100 instances were generated by the algorithm in each generated dataset, equally distributed between classes.

Similarly on the Breast Cancer Wisconsin diagnostic dataset, the mean accuracy of the neural networks was comparable between models trained on generated datasets, and models trained on training/testing batches (0.9377 and 0.9472 respectively, p = 0.1189).

|  | Generated data (large dataset) | Real world data (large dataset) | p value |
|---|---|---|---|
| Neural network mean accuracy [95% CI] | 0.956; [0.956-0.956] | 0.956; [0.9528-0.9605] | 0.6996 |
| Minimum accuracy | 0.956 | 0.95 | |
| Maximum accuracy | 0.956 | 0.96 | |

**Table 1A:** Performance of the algorithm on the Iris dataset – abundance of real world data

|  | Generated data (large dataset) | Real world data (large dataset) | p value |
|---|---|---|---|
| Neural network mean accuracy [95% CI] | 0.9377; [0.9185-0.9569] | 0.9472; [0.9278-0.9666] | 0.1189 |
| Minimum accuracy | 0.92 | 0.911 | |
| Maximum accuracy | 0.982 | 0.991 | |

**Table 1B:** Performance of the algorithm on the Breast Cancer Wisconsin diagnostic dataset – abundance of real world data

*3.4. Algorithm performance in conditions of simulated extreme scarcity of real world data*

At this point, the performance of the data generation algorithm was measured in conditions of simulated extreme scarcity of real world data. Fittest generated datasets obtained in simulated extreme scarcity of real world data, and corresponding scarce real world data training/testing batches were used to train neural networks. The performances of the machine learning models were comparatively analyzed on real world data validating batches. Overfitting was prevented as described earlier. Detailed results are shown in Table 2A and 2B.

*3.4.1. Performance on the Iris dataset*

For each experiment, the real world Iris dataset was randomly split into a scarce training/testing batch comprising only 1 instance per class (2% of total instances) and a validating batch (98% of instances). A total of 150 instances were generated in each generated dataset, equally distributed between classes. On the Iris dataset, the mean accuracy of the neural networks was significantly higher in models trained on generated datasets, compared to models trained on scarce training/testing batches (0.9533 and 0.9067 respectively, $p < 0.0001$).

*3.4.2. Performance on the Breast Cancer Wisconsin diagnostic dataset*

For each experiment, the real world Breast Cancer Wisconsin diagnostic dataset was randomly split into a scarce training/testing batch comprising only 1 instance per class (0.35% of total instances) and a validating batch (99.65% of instances). A total of 100 instances were generated in each generated dataset, equally distributed between classes.

On the Breast Cancer Wisconsin diagnostic dataset, the mean accuracy of the neural networks was significantly higher in models trained on generated datasets, compared to models trained on scarce training/testing batches (0.8692 and 0.7701 respectively, $p = 0.0091$). As a comparison, models trained on randomly generated datasets peaked at a mean accuracy of 0.4716.

|  | Generated data (on scarce dataset) | Real world data (scarce dataset) | p value |
|---|---|---|---|
| Neural network mean accuracy [95% CI] | 0.9533; [0.9395-0.9672] | 0.9067; [0.9028-0.9105] | <0.0001 *** |
| Minimum accuracy | 0.93 | 0.9 | |
| Maximum accuracy | 0.97 | 0.91 | |

**Table 2A:** Performance of the algorithm on the Iris dataset – scarcity of real world data

|  | Generated data (on scarce dataset) | Real world data (scarce dataset) | p value |
|---|---|---|---|
| Neural network mean accuracy [95% CI] | 0.8692; [0.833-0.9052] | 0.7701; [0.6722-0.8680] | 0.0091 ** |
| Minimum accuracy | 0.805 | 0.51 | |
| Maximum accuracy | 0.9115 | 0.867 | |

**Table 2B:** Performance of the algorithm on the Breast Cancer Wisconsin diagnostic dataset – scarcity of real world data

### 3.5. Distribution of generated datasets

Distributions of fittest generated datasets and real world datasets were analyzed on the Iris dataset.

*3.5.1. Distribution plot*

The comparative distribution of an actual fit generated dataset and real world dataset is shown in Figure 3, with instances and classes as a function of attributes.

*3.5.2. Distribution statistics*

The distributions were characterized in terms of central tendency (median and mean), and dispersion (25% and 75% percentiles). The analysis showed that the overall distributions were comparable. In more details, the distributions of the first, third and fourth attributes were not statistically different between generated dataset and real world dataset (with p values of 0.297, 0.3429, and 0.1553 respectively); the distributions of the second attribute significantly differed (p<0.0001), as shown in Table 3.

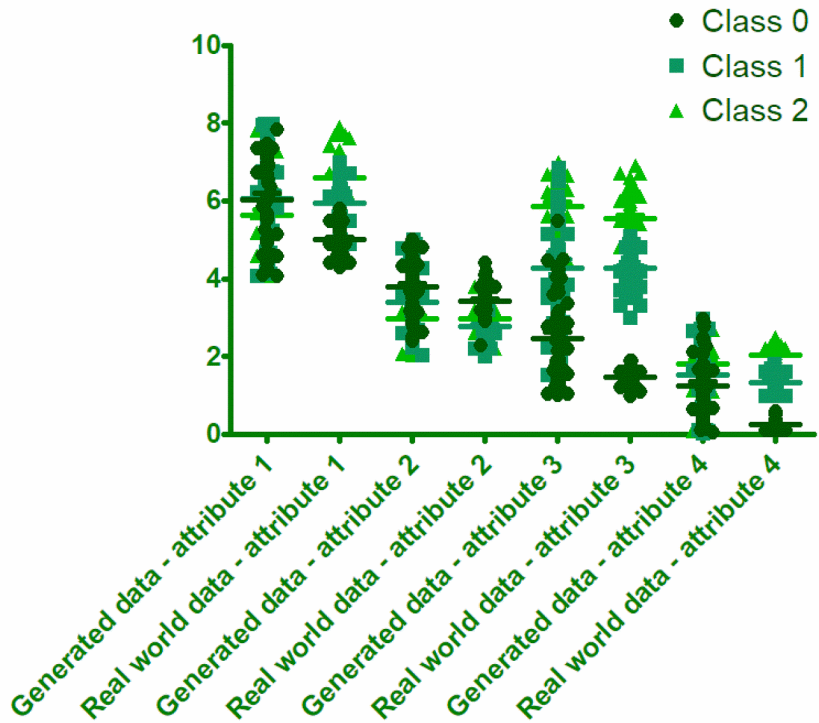

**Figure 3:** Distribution plot of fit generated dataset and real world dataset – Iris dataset

|  | Generated Data 1 | Real world Data 1 | Generated Data 2 | Real world Data 2 | Generated Data 3 | Real world Data 3 | Generated Data 4 | Real world Data 4 |
|---|---|---|---|---|---|---|---|---|
| Minimum | 4.023 | 4.3 | 2.01 | 2 | 1.007 | 1 | 0.0098 | 0.1 |
| 25% percentile | 5.062 | 5.1 | 2.727 | 2.8 | 1.987 | 1.575 | 0.5315 | 0.3 |
| Median | 6.051 | 5.8 | 3.47 | 3 | 3.435 | 4.35 | 1.231 | 1.3 |
| 75% percentile | 6.787 | 6.4 | 4.258 | 3.3 | 5.019 | 5.1 | 2.108 | 1.8 |
| Maximum | 7.998 | 7.9 | 4.91 | 4.4 | 6.928 | 6.9 | 2.963 | 2.5 |
| Mean | 5.979 | 5.843 | 3.469 | 3.054 | 3.594 | 3.759 | 1.343 | 1.199 |
| 95% CI | 5.801-6.158 | 5.710-5.977 | 3.336-3.602 | 2.984-3.124 | 3.305-3.883 | 3.474-4.043 | 1.203-1.483 | 1.076-1.322 |
| p value | 0.297 | | <0.0001 | | 0.3429 | | 0.1553 | |

**Table 3:** Distribution statistics of fit generated dataset and real world dataset – Iris dataset

## 3.6. Learning characteristics of the algorithm

The learning characteristics of the algorithm were analyzed on the Iris dataset.

The fitness of the fittest generated dataset progressively increases over the data generation process. This increase in fitness leads to an increase in the performance of the neural network used for data generation, and therefore to a decrease of its mean squared error.

### 3.6.1. Evolution of mean squared error during data generation

An example of the decrease of mean squared error during the data generation process is depicted in Figure 4.

As expected with most machine learning models, the learning curve of the algorithm is nonlinear, and shows an exponential regression type. Goodness of fit analysis shows an R-square value of 0.9441.

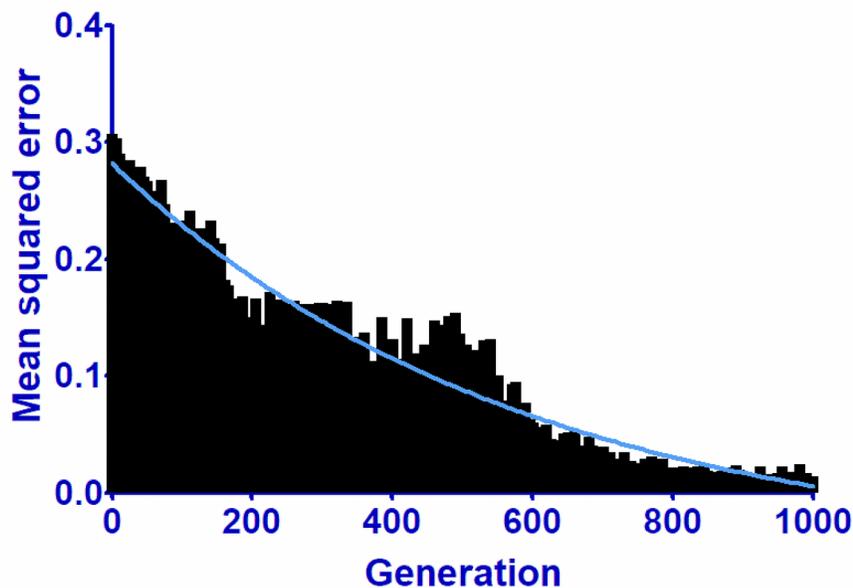

**Figure 4:** Evolution of mean squared error during the data generation process

*3.7. Algorithm time complexity and RAM usage*

The time complexity of the data generation algorithm was first theoretically derived, and then confirmed experimentally on the Iris dataset. RAM usage was measured experimentally on the Iris dataset.

*3.7.1. Theoretical time complexity*

First, theoretical complexity of genetic algorithms is known to be $O(g \times i \times n)$, with $g$ number of generations, $i$ number of individuals, $n$ size of individuals. Second, theoretical complexity of fully connected neural network is $O(e \times t(i \times j + j \times k))$, with $e$ number of epochs, $t$ number of training examples, $i$ number of nodes in input layer, $j$ number of nodes in hidden layer, $k$ number of nodes in output layer. Therefore, the theoretical time complexity of the data generation algorithm is expected to be $O(n^2)$.

*3.7.2. Experimental time complexity*

Time complexity of the algorithm was measured on the Iris dataset with respect to the number of inputs. The distributions of theoretical complexity and experimental complexity were not significantly different (p value of 0.2966), confirming that the algorithm is in practice of quadratic complexity with regards to the number of inputs, that is to say the amount of data to generate. Of note, goodness of fit analysis shows an R-square value of 0.9996.

*3.7.3. RAM usage*

RAM usage was measured with respect to the number of inputs. When the number of inputs doubles, RAM usage is increase by a factor of 1.02899. Similarly, when the number of inputs quadruples, RAM usage is increase by a factor of 1.0869. Therefore, RAM usage increases linearly with the number of generated data, for a fixed number of generations and a fixed number of candidate generated datasets per generation, which is confirmed by a goodness of fit analysis showing an R-square value of 1.

## 4. Discussion and conclusion

In this article, we have proposed a novel algorithm which can generate high quality, large artificial datasets to train machine learning models, even in conditions of extreme scarcity of real world data. The performance of machine learning models trained on generated data obtained in conditions of real world data abundance matches the performance of similar models trained on large real world datasets. Of note, this performance is also on par with previous descriptions of machine learning models trained on the Iris dataset and on the Breast Cancer Wisconsin diagnostic dataset reported in the literature [9, 10].

More importantly, the performance of machine learning models trained on generated data obtained in conditions of simulated extreme scarcity of real world data significantly exceeds the performance of similar models trained on scarce real world datasets. This observation is valid for all tested datasets, in the confirmed absence of overfitting.

It is noteworthy that data generated by the algorithm are not mathematically derived from real world data provided for the data generation process. Indeed, by definition, the neural network used in the data generation process will map the domain of the function it is trained to approximate with its co-domain. Since the neural network is trained with generated data and subsequently evaluated with real world data, and since selection pressure applied by the genetic algorithm ensures that the performance of the neural network progressively increases with both generated dataset and real world dataset, then, at the end of the data generation process, both datasets will be fit subsets of the domain and the co-domain of the same approximated function. But, by construction, generated data are not directly derived from available real world data. As a consequence, generated data are novel data elements, and not the products of a data augmentation procedure.

More precisely, the range of generated data is not limited to the range of real world data provided for the data generation process, but is extended to the entire domain of the random variable used by the genetic algorithm for applying point mutations. Therefore, if the domain of the random variable used by the algorithm matches the domain of the function to approximate instead of being restricted to the range of real world data available during the data generation process, then the algorithm will have a strong tendency to generalize.

A significant consequence of the previous observations is that the distribution of generated data may differ from the distribution of real world data provided for the data generation process. Indeed, as detailed above, by construction, generated dataset and real world dataset will be, once data generation process is complete, fit subsets of the domain and the co-domain of the same approximated function. This property does not necessarily imply that both datasets will significantly intersect; hence, their respective distributions may differ, as shown in this article (please refer to Figure 3 and Table 3). Interestingly, the data generation algorithm presented in this article also differs from previously described synthetic data generation algorithms. Indeed, as mentioned earlier, synthetic data generation algorithms need to be trained on a large amounts of real world data before they can start to generate accurate synthetic data. On the contrary, the data generation algorithm can generate large artificial datasets, even in conditions of extreme scarcity of real world data.

The complexity of the algorithm is also an import aspect to consider in the perspective of its deployment. Like many sorting algorithms, the data generation algorithm is of quadratic time complexity. This complexity allowed the algorithm to perform reasonably fast on the Iris dataset and on the Breast Cancer Wisconsin diagnostic dataset. Given the constantly increasing performance of computers and supercomputers, practical execution speed should not be a major issue, even on larger, more complex datasets.

To conclude, even if more experiments on more heterogeneous datasets are required to further validate the data generation algorithm presented in this article, the performance of the algorithm on the Iris dataset and on the Breast Cancer Wisconsin diagnostic dataset, reference datasets in machine learning experiments, leads us to believe that the data generation algorithm can successfully be used in real life situations of real world data scarcity, but also when cost or data sensitivity becomes an obstacle to the collection of large real world datasets.

# Acknowledgments

*Source code*

The source code of the data generation algorithm is protected by international intellectual property laws. It has been deposited at the Benelux Office for Intellectual Property under file numbers 138885 and 141030 (non-public deposits). Contact the author for commercial use.

*Article license*

This article is licensed under Creative Commons license CC BY-NC-SA 4.0: this license allows reusers to distribute, remix, adapt, and build upon the material in any medium or format for noncommercial purposes only, and only so long as attribution is given to the creator. If you remix, adapt, or build upon the material, you must license the modified material under identical terms.